\documentclass[conference]{IEEEtran}
\IEEEoverridecommandlockouts
\usepackage{times}
\usepackage{multirow}
\usepackage{cite}
\usepackage{amsmath,amssymb,amsfonts}
\usepackage{algorithmic}
\usepackage{graphicx}
\usepackage{textcomp}
\usepackage{xcolor}
\usepackage{cite}
\usepackage[numbers,sort&compress]{natbib}
\def\BibTeX{{\rm B\kern-.05em{\sc i\kern-.025em b}\kern-.08em
    T\kern-.1667em\lower.7ex\hbox{E}\kern-.125emX}}

\begin{document}
\title{LINK: Adaptive Modality Interaction for Audio-Visual Video Parsing\\


}
\title{LINK: Adaptive Modality Interaction for Audio-Visual Video Parsing}
\author{
	\IEEEauthorblockN{Langyu Wang$^{*1}$, Bingke Zhu\textsuperscript{\dag}$^2$, Yingying Chen$^2$, Jinqiao Wang$^2$}
	\IEEEauthorblockA{
		$^1$ School of Logistics Engineering, Shanghai Maritime University, China \\
		$^2$ Foundation Model Research Center, Institute of Automation, Chinese Academy of Sciences, China \\
        \IEEEauthorblockA{wangly54321@163.com, bingke.zhu@nlpr.ia.ac.cn, yingying.chen@nlpr.ia.ac.cn, jqwang@nlpr.ia.ac.cn \\
	\thanks{\dag Corresponding author. \newline * Word done as intern in CASIA.
	}
	}
        }
}
\maketitle

\begin{abstract}
Audio-visual video parsing focuses on classifying videos through weak labels while identifying events as either visible, audible, or both, alongside their respective temporal boundaries. Many methods ignore that different modalities often lack alignment, thereby introducing extra noise during modal interaction. In this work, we introduce a Learning Interaction method for Non-aligned Knowledge (LINK), designed to equilibrate the contributions of distinct modalities by dynamically adjusting their input during event prediction. Additionally, we leverage the semantic information of pseudo-labels as a priori knowledge to mitigate noise from other modalities. Our experimental findings demonstrate that our model outperforms existing methods on the LLP dataset.
\end{abstract}


\section{Introduction}
\begin{figure}                                                              
    \centering
    \includegraphics[height=5.0cm, width=\linewidth]{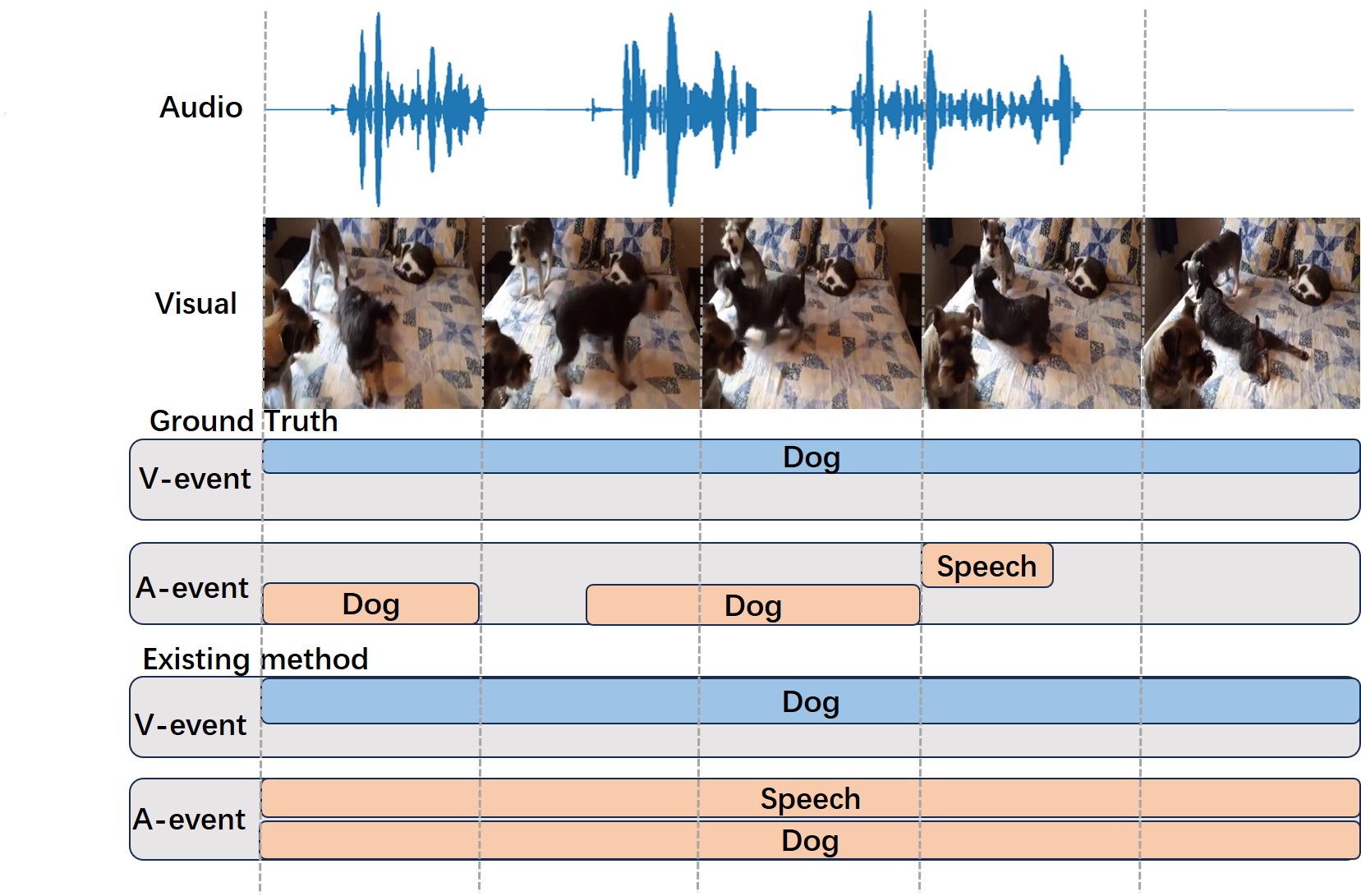}   
        \caption{\footnotesize Modality non-aligned samples from LLP. Existing method is vulnerable to non-aligned events and produce incorrect predictions.}
    \label{fig1}

\end{figure}

    As two pivotal components of multi-modal learning, visual and audio modalities play a pivotal role in comprehending diverse scenarios, including audio-visual event localization \cite{b1,b2}, audio-visual question answering \cite{b3}, and audio-visual action recognition \cite{b4}. Numerous researchers have assumed synchronization between visual and audio signals, positing that both modalities contain predictive clues for events \cite{b8}. However, in real-world scenarios, audio and visual events are not always aligned. In other words, different modalities may not always provide useful cues for predicting events. As illustrated in Fig. \ref{fig1}, even though several dogs are present in the scene, the corresponding audio ground truth predominantly features speech rather than the dogs' sounds, highlighting the misalignment between audio and visuals.

    Observing the misalignment of modalities in generic videos, Tian et al \cite{b5} introduced the Audio-Visual Video Parsing (AVVP) task. As depicted in Fig. \ref{fig1}, AVVP leverages both audio and visual inputs to identify event categories, modalities, and temporal boundaries. Two primary strategies for this task include enhancing model architectures and refining labeling methods. Architectural enhancements include combination of Hybrid Attention Networks (HAN) \cite{b5}, Multi-Modal Pyramidal Feature Attention \cite{b6}, and Dual-Guided Attention \cite{b7}. Furthermore, the advent of large-scale pre-training models has refined labeling by providing more detailed supervision. For instance, Lai et al. \cite{b8} employed frozen CLIP and CLAP over traditional Resnet and VGGish for feature extraction and generation of dense pseudo-labels to assist in predictions. Following this, Zhou et al. \cite{b9} proposed a novel decoding strategy to address the parsing of potentially overlapping events. Despite these advancements, existing methods fail to account for the differential impacts of modalities on event prediction during their interaction, leading to decreased prediction accuracy due to the incorporation of irrelevant noise when audio and visual events are non-aligned.

    To address these challenges, we introduce a Learning Interaction method for Non-aligned Knowledge (LINK), designed to equilibrate the contributions of disparate modalities. This strategy dynamically modulates the interaction weights between the visual and audio channels, employing the semantic content of pseudo-labels to guide the prediction process of each modality. Initially, our method includes a loss function predicated on segmental weights, utilizing cosine similarity between modes to variably weight samples across different ranges. Subsequently, we leverage temporal-spatial attention for modality-specific enhancements and facilitate cross-modal synthesis through adaptively weighted interactions, aiming to harmonize modal input. Lastly, we incorporate uni-modal pseudo-labels derived from CLIP and CLAP as constraints within the model, thereby refining the accuracy of predictions.

\begin{figure*}
    \centering
    \setlength{\abovecaptionskip}{-0.1cm}
    \includegraphics[height=7.5cm, width=\linewidth]{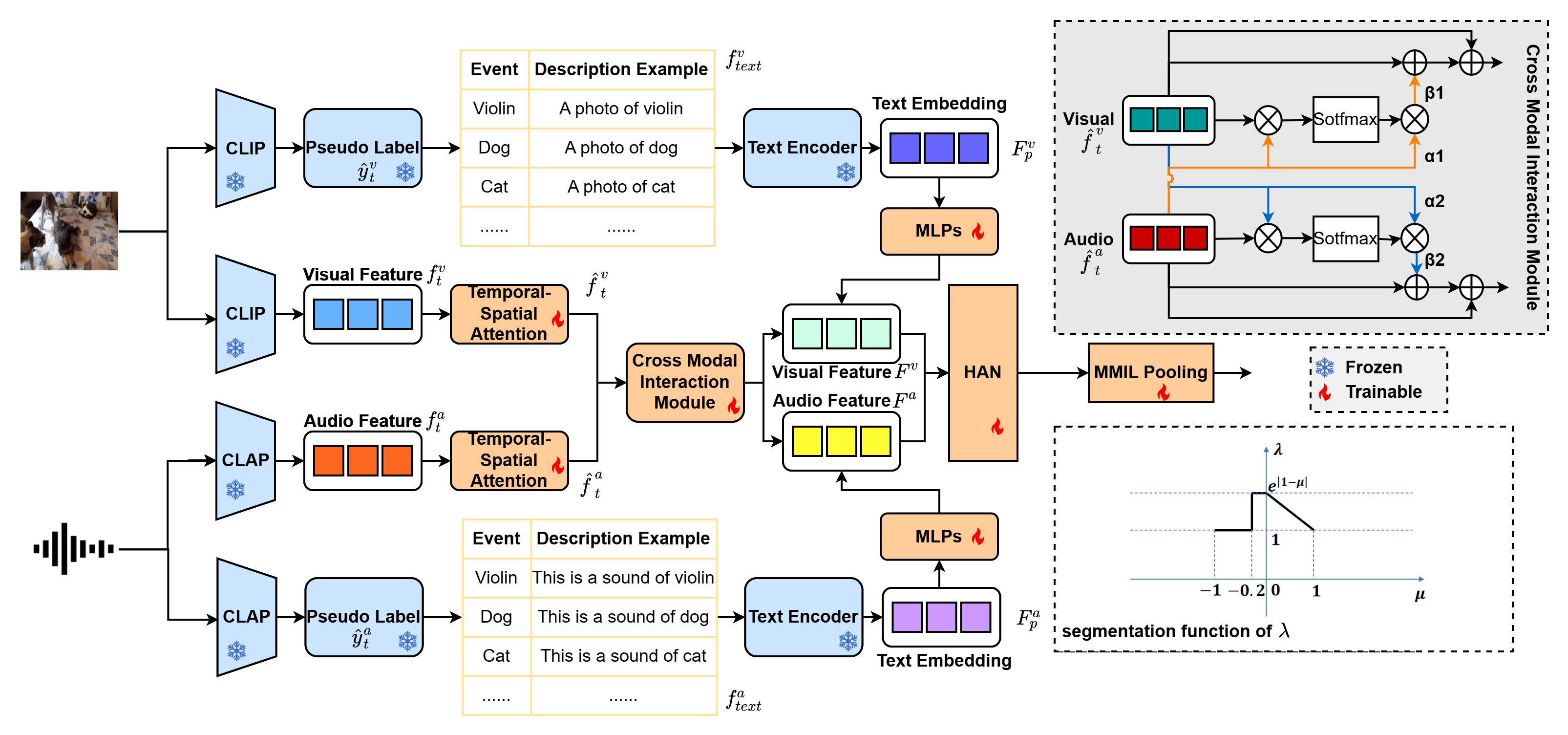}
    \caption{\footnotesize The framework of \textbf{LINK}. We use temporal-spatial attention and cross modal interaction module to enhance the expression of feature, and merge the semantic information from pseudo label with uni-modal feature. The pseudo labels are extracted by VALOR\cite{b8}.}\vspace{-2pt}
    \label{fig:enter-label}
      
\end{figure*}

\section{PROPOSED METHOD}

\subsection{Problem Statement}\label{AA}
The AVVP task aims to identify the event of every segment into audio event, visual event and audio-visual event, together with their classes. For the benchmark dataset of Look, Listen, and Parse (LLP), a T-second video is split into T non-overlapping segments, expressed as S = $\lbrace A_t,V_t\rbrace ^T_{t=1}$, where A and V represent audio and visual segment in time t respectively. In each segment, $y^a_t \in \mathbb R^C$, $y^v_t \in \mathbb R^C$, $y^{av}_t \in \mathbb R^C$ represent to the audio event labels, visual event labels and audio-visual event labels, $C$ is the number of event types. However, we only have weak labels in training split, but have detailed event labels with modalities and temporal boundaries for evaluation.

\subsection{Temporal-Spatial Attention and Adaptive Modality Interaction Module (TSAM)}
As shown in Fig.2, we obtain audio and visual features by using pre-trained audio encoder CLAP and visual encoder CLIP \cite{b8}, denoted as $\lbrace f^a_t \rbrace ^T_{t=1}$, $\lbrace f^v_t \rbrace ^T_{t=1}$. $F^a$ and $F^v$ stand for the feature set in the same video, which are defined as $F^a =\lbrace f^a_1, ..., f^a_T \rbrace \in \mathbb R^{T\times d}$ and $F^v =\lbrace f^v_1, ..., f^v_T \rbrace \in \mathbb R^{T\times d}$, d is the feature dimension. In order to aggregate features from different segments and improve the feature expressiveness of each segment, we propose a segment-based temporal-spatial attention module, which is similar to the convolutional block attention module \cite{b10}. 

 \textbf{Temporal attention.} We generate two feature vectors by global maximum pooling and global average pooling for each segment. After the feature vectors are fed into a shared fully connected layer, we can obtain the final attention weight vector $W^m_t$, $m \in \{a,v\}$:
\begin{equation}
\resizebox{0.9\hsize}{!}{$
    W^m_t(f^m_t) = \delta (MLP(AvgPool(f^m_t))+MLP(MaxPool(f^m_t ))),
$}
\end{equation}
where $f^m_t$ denotes audio or visual features within a video, whose dimensions are $(B, T, D)$, representing batchsize, segments, and feture, respectively. $\delta$ indicates sigmoid function.

\textbf{Spatial attention.} Similar to temporal attention, we generate two feature vectors by max pooling and average pooling along the temporal (segment) dimension. The features after max pooling and average pooling are then concatenated along the temporal dimension to obtain a feature vector, which is referred to as spatial attention $S^m_t$, $m \in \{a,v\}$ :
\begin{equation}
\label{eq3}
    S^m_t(f^m_t) = \delta ((AvgPool(f^m_t));(MaxPool(f^m_t ))).     
\end{equation}

The outputs of the channel attention module and the spatial attention module are element-wise multiplied to obtain the final attention-enhanced feature $\hat{f}^m_t$, $m \in \{a,v\}$:
\begin{equation}
    \hat{f}^m_t =S^m_t(W^m_t(f^m_t) \otimes f^m_t)\cdot (W^m_t(f^m_t) \cdot f^a_t).
\end{equation}

Subsequently, these features are input into the cross modal interaction module (CMIM). We have incorporated the AV-Adapter \cite{b11} from recent work to enable adaptive cross-modal interaction. We use four learnable parameters to balance the contributions of different modalities in the modal fusion. The whole produce of CMIM can be expressed as:
\begin{equation}
    \hat{f}^{ac}_t = \hat{f}^a_t + \alpha_2 \cdot Softmax(\hat{f}^a_t \hat{f}^v_t )(\beta_2 \cdot \hat{f}^v_t),
\end{equation}
\begin{equation}
    \hat{f}^{vc}_t = \hat{f}^v_t + \alpha_1 \cdot Softmax(\hat{f}^v_t \hat{f}^a_t) (\beta_1 \cdot \hat{f}^a_t),
\end{equation}
\begin{equation}
    \hat{f}^{aout}_t = \hat{f}^a_t + \hat{f}^{ac}_t , \hat{f}^{vout}_t = \hat{f}^v_t + \hat{f}^{vc}_t,
\end{equation}
where $\alpha_1$ ,$\alpha_2$, $\beta_1$ and $\beta_2$ are the trainable weights, $\hat{f}^{aout}_t$ and $\hat{f}^{vout}_t$ are the final outputs of CMIM.
\subsection{Segmented Audio-Visual Semantic Similarity Loss Function (S-LOSS)}
In order to more efficiently optimise the model after performing feature fusion, we combine the pseudo-label usage as well as the loss function proposed in recent work \cite{b8,b9}. The loss function can be expressed as: 
\begin{equation}
    \mathcal L_{VALOR} = \mathcal L_{video} + \mathcal L^a_{video} + \mathcal L^v_{video} ,
\end{equation}
\begin{equation}
    \mathcal L_{video} = BCE(p,y), \mathcal L^m_{video} = \sum_t BCE(p^m_t,\hat{y}^{mgt}_t) ,
\end{equation}
where $p^m_t$ denotes the audio and visual event probabilities, $\hat{y}^{mgt}_t$ denotes audio and visual pseudo-labels, $p$ denotes audio-visual event probabilities, $y$ is the truth video-level labels.

We incorporate the segment-wise pseudo labels generated in recent work \cite{b8, b12, b13} to provide fine-grained supervision. The video-level pseudo-labels $Y$ are also easy to obtain, and the event will be included in the video-level pseudo-labels as long as the event occurs in the temporal segments. Thus the loss function can be rewritten as:
\begin{equation}
    \mathcal L_{VALOR} = \mathcal L_{video} + \mathcal L^a_{video} + \mathcal L^v_{video} + \mathcal L_{label} + \mathcal L_{avss} ,
\end{equation}
\begin{equation}
    \mathcal L_{label} = BCE(Y,y), \mathcal L_{avss} = MSE(s,r) ,
\end{equation}
where $s$ denotes the the cosine similarity of all audio-visual segment pairs, $r$ represents intersection over union of audio events and visual events, which ranges between [0,1]. 

However, the prediction results of multi-modal models are not necessarily better than those of uni-modal models \cite{b14, b15}. Even at the same frame, audio and video are likely to be non-aligned, thus negatively affecting the prediction results of the multi-modal model \cite{b16}. In fact, a significant portion of this mismatched data is likely to become noise, thus affecting the prediction of events in modal interactions.

 To optimise the training process as much as possible, we use the cosine similarity as the basis for the matching degree of visual and audio features. When the cosine similarity of visual and audio features is closer to 1, it indicates that the two events are more similar, and the modal interaction is facilitates the improvement of the multi-modal model prediction; when it is closer to $-1$, it indicates that the two events are less similar, and the modal interaction limits the multi-modal model prediction. We reduce the hyperparameters by directly using the cosine similarity equal to 0 as a threshold. In order for the model to better optimise those samples that are at the boundary and to make fuller use of these samples to train the network, we try to give a larger weight to the samples in the part of the cosine similarity in the range of ($-0.2$, 0) and hope that the samples in this part of the range will dominate the loss function as much as possible. At the same time, a decreasing weight is assigned to the samples with cosine similarity in (0, 1) and a smaller weight is assigned to the samples with cosine similarity in ($-1$, $-0.2$). The function can be expressed as:
\begin{equation}
\lambda = 
\begin{cases}
    1& \quad s\le -0.2\\
    e^{|1-\mu|}& \quad -0.2<s<0\\
    e^{|1-\mu|}+(1-e^{|1-\mu|})s& \quad s \ge 0
\end{cases}
\end{equation}
where $\mu$ represents a trainable parameter. The final loss function can be expressed as:
\begin{equation}
    \mathcal L_{VALOR} = \mathcal L_{video} + \mathcal L^a_{video} + \mathcal L^v_{video} + \mathcal L_{label} +\lambda \mathcal L_{avss} .
\end{equation}
\subsection{Pseudo Label Semantic Interaction Module (PLSIM)}
The information in language can be used as a cue to effectively improve the performance of the model \cite{b9, b17, b18, b25}. Liu et al. \cite{b24} proposed a vision-audio-language pretraining model, which can learn strong multi-modal correlations and generalize to various downstream tasks. In AVVP task, pseudo-labels are proposed to assist in model prediction, and better results have been achieved after introducing pseudo-labels into uni-modal loss. However, this is not the best use of pseudo-labels. The semantic information of uni-modal pseudo-labels is not sufficiently integrated with the features to achieve the effect of constraining uni-modal event prediction. The pseudo-labels can serve as a priori information to culling out the other modal noise. Unlike the pseudo labels $\hat{y}^{mgt}_t$ \cite{b8} used in the loss function, we removed the logical operations with the video-level labels when generating pseudo labels for testing purposes. New pseudo labels can be expressed as $\hat{y}^m_t$.

\begin{table*}[]
\centering
\caption{\footnotesize Comparison with the state-of-the-art methods on the LLP dataset in terms of F-scores.}
\label{tab:my-table}
\begin{tabular}{c|c|ccccc|ccccc}
\hline
                                      &                                   & \multicolumn{5}{c|}{\textbf{Segment-level}}                                                                                                         & \multicolumn{5}{c}{\textbf{Event-level}}                                                                                                            \\ \cline{3-12} 
\multirow{-2}{*}{\textbf{Method}}     & \multirow{-2}{*}{\textbf{Venue}}  & \textbf{A}                  & \textbf{V}                  & \textbf{AV}                 & \textbf{Type@AV}            & \textbf{Event@AV}           & \textbf{A}                  & \textbf{V}                  & \textbf{AV}                 & \textbf{Type@AV}            & \textbf{Event@AV}           \\ \hline
HAN\cite{b5}                             & ECCV’20                           & 60.1                        & 52.9                        & 48.9                        & 54.0                        & 55.4                        & 51.3                        & 48.9                        & 43.0                        & 47.7                        & 48.0                        \\
MM-Pyr\cite{b6}                          & MM’22                             & 60.9                        & 54.4                        & 50.0                        & 55.1                        & 57.6                        & 52.7                        & 51.8                        & 44.4                        & 49.9                        & 50.0                        \\
MGN\cite{b19}                           & NeurIPS’22                        & 60.8                        & 55.4                        & 50.4                        & 55.5                        & 57.2                        & 51.1                        & 52.4                        & 44.4                        & 49.3                        & 49.1                        \\
JoMoLD\cite{b20}                        & ECCV’22                           & 61.3                        & 63.8                        & 57.2                        & 60.8                        & 59.9                        & 53.9                        & 59.9                        & 49.6                        & 54.5                        & 52.5                        \\
CMPAE\cite{b21}                         & CVPR’23                           & 64.2                        & 66.4                        & 59.2                        & 63.3                        & 62.8                        & 56.6                        & 63.7                        & 51.8                        & 57.4                        & 55.7                        \\
DGSCT\cite{b7}                          & NeurIPS’23                        & 59.0                        & 59.4                        & 52.8                        & 57.1                        & 57.0                        & 49.2                        & 56.1                        & 46.1                        & 50.5                        & 49.1                        \\
{\color[HTML]{333333} VALOR++\cite{b8}} & {\color[HTML]{333333} NeurIPS’23} & {\color[HTML]{333333} 68.1} & {\color[HTML]{333333} 68.4} & {\color[HTML]{333333} 61.9} & {\color[HTML]{333333} 66.2} & {\color[HTML]{333333} 66.8} & {\color[HTML]{333333} 61.2} & {\color[HTML]{333333} 64.7} & {\color[HTML]{333333} 55.5} & {\color[HTML]{333333} 60.4} & {\color[HTML]{333333} 59.0} \\
CM-PIE\cite{b23}                         & ICASSP’24          & 61.7                       & 55.2                        & 50.1                        & 55.7                       & 56.8                        & 53.7                        & 51.3                        & 43.6                       & 49.5                        & 51.3                        \\
LEAP\cite{b9}                         & ECCV’24                           & 64.8                        & 67.7                        & 61.8                        & 64.8                        & 63.6                        & 59.2                        & 64.9                        & 56.5                        & 60.2                        & 57.4                        \\
CoLeaF+\cite{b22}                       & ECCV’24                           & 64.2                        & 67.1                        & 59.8                        & 63.8                        & 61.9                        & 57.1                        & 64.8                        & 52.8                        & 58.2                        & 55.5                        \\
 \hline
\textbf{LINK (Ours)}                         & \textbf{-}                        & \textbf{69.7}               & \textbf{69.0}               & \textbf{62.1}               & \textbf{66.9}               & \textbf{68.5}               & \textbf{63.4}               & \textbf{64.9}               & \textbf{55.7}               & \textbf{61.3}               & \textbf{60.8}               \\ \hline
\end{tabular}
\end{table*}
Since both pseudo labels $\hat{y}^m_t$ and real labels $y$ are one-hot coded, the event categories corresponding to pseudo-labels $f^a_{text}$ and $f^v_{text}$ can be easily derived from the coding of pseudo-labels. Next, we convert the event categories in the pseudo-labels into concepts that can be understood by CLIP and CLAP. The title of each event is constructed by adding the prefix ‘A photo of’ or 'this is a sound of' to the natural language form of the event. These captions are processed by a frozen CLIP/CLAP text encoder to obtain pseudo label semantic features $F^a_{p}$ and $F^v_{p}$ for good linguistic consistency: 
\begin{equation}
    F^a_p = CLAP(f^a_{text}), F^v_p = CLIP(f^v_{text}).
\end{equation}
 
 In addition, we use multiple MLPs $\varDelta^n_m$ to map the semantic information of the text, which can be represented as:
\begin{equation}
    \gamma_{a1} = \varDelta^1_m(F^a_p),\gamma_{a2} = \varDelta^2_m(F^a_p),
\end{equation}
\begin{equation}
    \rho_{v1} = \varDelta^3_m(F^v_p),  \rho_{v2} = \varDelta^4_m(F^v_p),
\end{equation}
where $\varDelta^1_m$, $\varDelta^2_m$, $\varDelta^3_m$, $\varDelta^4_m$ are different MLPs operations to generate the semantic parameters respectively. We use the extracted audio/visual features to fuse with the semantic information, which can be represented as:
\begin{equation}
    F_a = \hat{f}^{aout}_t \odot \gamma_{a1} + \gamma_{a2} + \hat{f}^{aout}_t,
\end{equation}
\begin{equation}
    F_v = \hat{f}^{vout}_t \odot \rho_{v1} + \rho_{v2}+ \hat{f}^{vout}_t,
\end{equation}
where $\odot$ denotes Hadamard product. $\gamma_{a1}$ and $\rho_{v1}$ denote scale scaling, $\gamma_{a2}$ and $\rho_{v2}$ denote bias control. $F_a$ and $F_v$ are audio features and visual features fused with semantic features.

\section{EXPERIMENTAL RESULTS}

\subsection{Experimental setup}\label{AA}
\textbf{Datasets.} The LLP dataset \cite{b5} is used to evaluate our method.
This dataset has 11849 videos with 25 categories taken from YouTube, containing various scenes and species. The dataset has 10000 videos with weak labels as the training set, 1200 videos and 649 videos as the testing set and the validation set with fully annotated labels. 


\subsection{Comparison with State-of-the-art Methods}\label{AA}
We compare our method with several popular baselines, such as HAN \cite{b5}, MM-Pyramid \cite{b6}, VALOR++ \cite{b18}, DG-SCT \cite{b7}. Table 1 shows the results that our method outperforms the benchmark models VALOR++ and HAN in all the metrics, and compared with VALOR++, it achieves improvements in uni-modal performance, e.g., 1.6\% at the audio segment level (69.7\% vs. 68.1\%) and 2.2\% at the audio event level (63.4\% vs. 61.2\%). Meanwhile, the multi-modal performance is also improved, e.g., 0.2\% at the AV segment level (62.1\% vs. 61.9\%) and 0.2\% at the AV event level (55.7\% vs. 55.5\%).

We also achieve the state-of-the-art (SOTA) compared to other models. Compared with \cite{b6, b7, b19, b20, b21, b22}, we use CLIP and CLAP instead of Resnet and VGGish to extract visual and audio features, and there is a big breakthrough in the expression ability of features. Compared with [8,9], which use CLIP and CLAP, we better balance the contribution of different modalities in fusion and effectively suppress modal noise, which makes the model achieve satisfactory results in all metrics, especially in the prediction of uni-modal events.	

\begin{table}[]
\centering
\caption{\footnotesize Ablation Study on the LLP. w/ denotes adding the component.}
\label{tab:my-table}
\begin{tabular}{cccccc}
\hline
\multicolumn{6}{c}{\textbf{Segment-level}}                                                     \\ \hline
\textbf{Method} & \textbf{A} & \textbf{V} & \textbf{AV} & \textbf{Type@AV} & \textbf{Event@AV} \\ \hline
VALOR++         & 68.1       & 68.4       & 61.9        & 66.2             & 66.8              \\
w/ TSAM(T)      & 68.8       & 68.7       & 61.7        & 66.4             & 67.4              \\
w/ S-LOSS(S)    & 68.4       & 68.3       & 62.0        & 66.2             & 67.6              \\ 
w/ PLSIM(P)     & 69.0       & 68.2       & 61.3        & 66.2             & 68.1              \\ \hline
w/ T and S      & 69.5       & 68.8       & 62.0        & 66.8             & 68.0              \\
w/ S and P      & 69.3       & 68.6       & 61.7        & 66.7             & 68.1              \\
w/ T and P      & 69.3       & 68.6       & 62.1        & 66.7             & 68.4              \\ \hline
\multicolumn{6}{c}{\textbf{Event-level}}                                                       \\ \hline
\textbf{Method} & \textbf{A} & \textbf{V} & \textbf{AV} & \textbf{Type@AV} & \textbf{Event@AV} \\ \hline
VALOR++         & 61.2       & 64.7       & 55.5        & 60.4             & 59.0              \\ 
w/ TSAM(T)      & 62.0       & 63.4       & 55.1        & 60.2             & 60.2              \\
w/ S-LOSS(S)    & 62.7       & 64.4       & 55.5        & 60.9             & 59.9              \\
w/ PLSIM(P)     & 62.9       & 63.6       & 55.1        & 60.5             & 60.2              \\ \hline
w/ T and S      & 63.1      & 63.7        & 55.1        & 60.6             & 60.5              \\
w/ S and P      & 63.2       & 64.4       & 55.5        & 61.0             & 60.5              \\
w/ T and P      & 63.1       & 64.7       & 55.5        & 61.1             & 60.3              \\ \hline          
\end{tabular}
\end{table}

\subsection{Ablation Study}                       

\textbf{Ablation experiments on the training strategy (loss function).} The segment weight-based loss function aims to optimise those samples with similarity between ($-0.2$, 0) to dominate the loss function, trying to eliminate the negative impact of dissimilar events in modal fusion. 
The results show that the loss function improves on six metrics such as audio segment-level, visual segment-level, and especially the improvement on uni-modal event prediction is more obvious. These results show that the segment weight-based loss function can better correct the interaction effects of different modalities while quantifying the correlation of various modalities.

\textbf{Ablation experiments on TSAM.} The TSAM module captures the fine-grained features of each segment, to selectively enhance or weaken the importance of different segments, while doing so with attention to time and space. Meanwhile, the four trainable parameters in the CMIM allow the model to selectively enhance or suppress the influence of another modality when performing cross-modal interactions, thus balancing the modal contributions.

\textbf{Ablation experiments on the PLSIM.} PLSIM extracts event categories based on pseudo-labels and converts them into textual information that can be recognised by CLIP. It is fed into a pre-trained text encoder to get the speech information, which is then fused with the features of the Modal Interaction Module. The semantic information of the pseudo-labels can be used as a kind of a priori information, greatly eliminating the noise effect of another modality and improving the prediction ability of the model in uni-modal events. 

\section{CONCLUSION} 
In this paper, we have presented a novel weakly-supervised audio-visual video parsing framework LINK. We use a training strategy with a loss function of segmented weights and weighted interactions after temporal-spatial attention to the extracted features to balance the modal contributions. In addition, we use a pseudo labels semantic information fusion module that greatly improves the prediction of uni-modal events. In future work, we will further investigate how to improve the prediction of multi-modal events.

\end{document}